\documentclass{article}
\usepackage{amsmath,amsthm,amssymb,bbm} 
\usepackage{amsfonts}
\usepackage[preprint]{corl_2022} % Uncomment for pre-prints (e.g., arxiv); This is like ``final'', but will remove the CORL footnote.

\DeclareMathOperator*{\argmin}{arg\,min}

\theoremstyle{definition}

\newtheorem{definition}{Definition}

\usepackage{xspace}

\newcommand{\karen}[1]{}
\newcommand{\aw}[1]{}
\newcommand{\mg}[1]{}
{}       % Karen comments

\usepackage{algorithmicx} %
\usepackage{algorithm}%
\usepackage[noend]{algpseudocode}%
\usepackage{bm}
\usepackage{wrapfig}

\usepackage{booktabs}
\usepackage{graphicx}
\usepackage{subcaption}
\usepackage{makecell} 

\usepackage[small,compact]{titlesec}
\titlespacing*{\subsubsection}{0pt}{4pt plus 2pt minus 2pt}{0pt plus 1pt minus 1pt}

\usepackage{enumitem}
\setlist[itemize]{itemsep=0pt, topsep=0pt, leftmargin=25pt}
\setlist[enumerate]{itemsep=0pt, topsep=0pt, leftmargin=14pt}

\usepackage[title,toc,titletoc,page]{appendix} %appendix
\algrenewcommand\algorithmicrequire{\textbf{Input:}}
\algrenewcommand\algorithmicensure{\textbf{Output:}}

\title{Learning Diverse and Physically Feasible\\ Dexterous Grasps with Generative Model and \\
Bilevel Optimization}

\author{
  Albert Wu\\
  Computer Science Department\\
  Stanford University,
  United States\\
  \texttt{amhwu@stanford.edu} \\
  \And
  Michelle Guo\\
  Computer Science Department\\
  Stanford University,
  United States\\
  \texttt{mguo95@cs.stanford.edu} \\
  \And
  C. Karen Liu\\
  Computer Science Department\\
  Stanford University,
  United States\\
  \texttt{karenliu@cs.stanford.edu} \\
}

\begin{document}
\maketitle

%===============================================================================

\begin{abstract}
% Problem
To fully utilize the versatility of a multi-fingered dexterous robotic hand for executing diverse object grasps, one must consider the rich physical constraints introduced by hand-object interaction and object geometry.
We propose an integrative approach of combining a generative model and a bilevel optimization (BO) to plan diverse grasp configurations on novel objects. First, a conditional variational autoencoder trained on merely six YCB objects predicts the finger placement directly from the object point cloud. The prediction is then used to seed a nonconvex BO that solves for a grasp configuration under collision, reachability, wrench closure, and friction constraints.
Our method achieved an $86.7\%$ success over $120$ real world grasping trials on $20$ household objects, including unseen and challenging geometries.
Through quantitative empirical evaluations, we confirm that grasp configurations produced by our pipeline are indeed guaranteed to satisfy kinematic and dynamic constraints. 
A video summary of our results is available at \url{youtu.be/9DTrImbN99I}.

\end{abstract}

% Two or three meaningful keywords should be added here

%===============================================================================

\section{Introduction}
\label{sec:introduction}
% Why is it interesting and important?
% What is the problem
% Dexterous grasping is a fundamental problem in robotics
Performing diverse grasps on a variety of objects is a fundamental skill in robotic manipulation. Diverse grasp configurations allow flexible interaction with the objects while satisfying requirements demanded by the higher level task of interest.
\textit{Dexterous grasping}, which refers to object grasping with a fully actuated, multi-finger dexterous robotic hand, has the potential to achieve diverse grasp configurations with applicability to a wide range of objects. 
This is in contrast with \textit{simple grasps} achieved by parallel jaw grippers or underactuated multi-finger grippers, both of which have fingertip workspaces restricted to a subspace of the 3D task space.

% Why is it hard? 
% Identify the two
We identify two major challenges in planning diverse dexterous grasps. Firstly, the solution space is multimodal due to the many finger placement possibilities and the lack of a metric defining an optimal grasp among valid answers. Secondly, each valid grasp is governed by nonconvex physical constraints such as collision, contact, and force closure.

The multimodality of dexterous grasp planning motivates the use of deep learning-based approaches. It is challenging to produce diverse and multimodal grasp plans with regression or naive supervised methods. Analytical approaches such as precomputing a grasp library or sampling-based grasp planning, while popular for planning simple grasps, are computationally intractable on high-dimensional dexterous grasps. Nevertheless, learning to plan grasps that strictly obey physical constraints is difficult.
Model-based optimization has historically been applied to enforce exact constraints. However, in practice the dexterous grasp planning problem is arguably too nonconvex to solve directly. Consequently, previous works usually relax the constraints as scalar losses and solve the relaxed problem, sacrificing exact constraint satisfaction guarantees for practicality.

We propose an integrative approach that combines learning and optimization to produce diverse physically-feasible dexterous grasp configurations for unseen objects. Our method first predicts an initial set of contact points using a conditional variational autoencoder (CVAE). The contact points are then projected onto the manifold of kinematically and dynamically feasible grasps by solving a bilevel optimization (BO) problem. Our key contributions are summarized below:\\
    $\bullet$ \textbf{Learning-based dexterous grasp planning pipeline} that integrates CVAE and BO to produce diverse, fully-specified, and constraint-satisfying dexterous grasps from point clouds. \\
    % $\bullet$ \textbf{CVAE-based dexterous grasp prediction network} that takes in an object point cloud and produces diverse grasp finger placement predictions parameterized by the latent variable sample.\\
    $\bullet$ \textbf{Bilevel grasp optimization formulation} that takes a learning-based grasp prediction and outputs a dexterous grasp that satisfies reachability, collision, wrench closure, and friction cone constraints.\\
    $\bullet$ \textbf{Successful and extensive hardware validation} on 20 household objects. Our method achieved an overall success rate of $86.7\%$ over $120$ grasp trials.\\
    % \textbf{Empirical evidence on integrating learning and model-based methods}, obtained though metrics inspired by physics.

\begin{figure}[!htb]
    \centering
    \includegraphics[width=\textwidth]{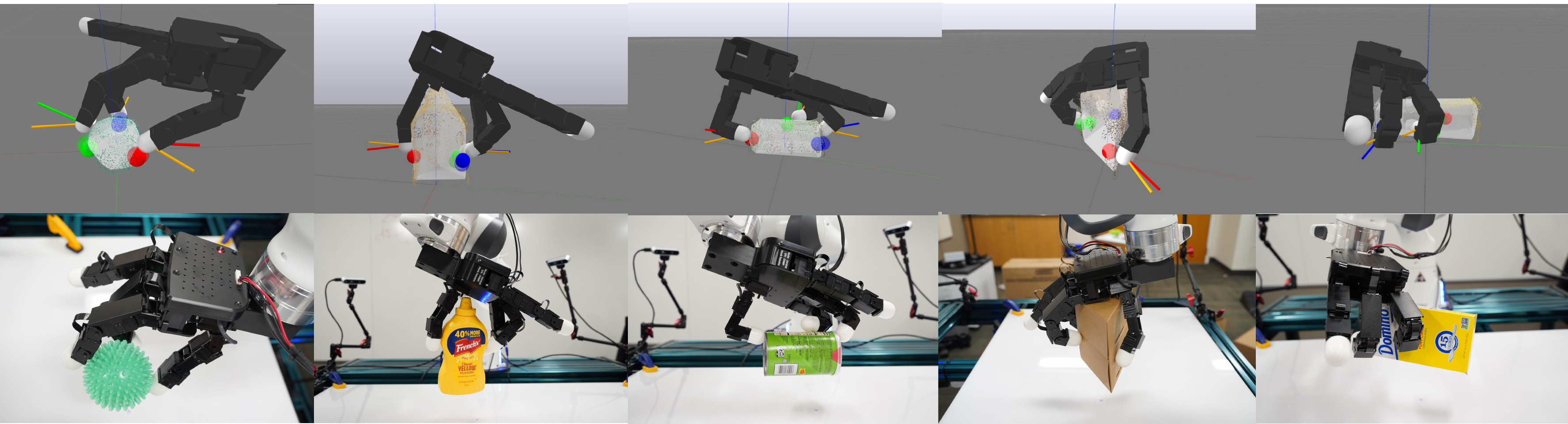}
    \caption{Our method can pick up different objects shapes with a diverse set of grasp configurations. }
    \label{fig:teaser}
\end{figure}

\section{Related Work}
\label{sec:related_work}
We limit our discussion to literature on grasping in uncluttered scenes, with an emphasis on dexterous grasping. We exclude work that focuses on object segmentation in cluttered environments or other types of manipulation, such as in-hand manipulation. 
% Robot manipulation has been an active and broad research area for decades. We limit our discussion to papers that focus on grasping in uncluttered scenes. We exclude work that focuses on object segmentation in cluttered environments or other types of manipulation, such as in-hand manipulation. 

% \subsection{Simple Grasping with Wrist Pose Prediction}
\subsection{Dexterous Grasping}
% \subsection{Learning in Dexterous Grasping}
Dexterous grasping is a long standing problem in the robotics community. In general, the literature can be split into learning-based and analytical methods.
Early learning-based grasp planners seek to fit the space of possible grasps for rapid grasp generation (e.g.,~\citep{huang2013learning}). More recent papers have shifted to producing grasps directly with complex model architectures, such as generative models~\citep{lundell2021ddgc,shao2020unigrasp, wei2022dvgg} and convolutional neural networks~\citep{lu2020planning,mandikal2021learning}. However, most of these works do not account for physical laws and have restricted, or even missing, hardware evaluation. This issue is exacerbated by the difficulty of simulating contact-rich interactions \citep{pfrommer2020contactnets, jiang2022data}. The quality of grasps produced by learning exclusively with simulation is questionable. Moreover, many papers do not emphasize the ability to learn \textit{diverse} grasps, thus forgoing a key benefit of dexterous hands over simple grippers. Notably, some publications on learning dexterous grasping originate from computer vision and graphics communities (e.g.,~\citep{zhu2021toward, brahmbhatt2020contactpose, taheri2020grab, jiang2021hand, yang2021cpf, corona2020ganhand}). While these papers cover diverse grasp generation on fully dexterous hands, their ultimate objective tends to be achieving photorealistic human grasps in simulation rather than satisfying strict real-world physical constraints.

% \subsection{Analytical Techniques in Dexterous Grasping}
On the other hand, analytical dexterous grasp planners are often inspired by physical constraints such as force closure, friction cone, robustness to external disturbance wrenches, and object contact (e.g.,~\cite{ciocarlie2009hand, pokorny2013classical, ferrari1992planning,liu2021synthesizing}). Historically, analytical methods are standalone and includes both grasp modality exploration and physical constraint compliance. We point the readers to~\cite{roa2015grasp, bohg2013data} for a detailed review on these methods. Some limitations of these approaches include dependency on a high-fidelity object model and lack of diversity in the generated grasps~\citep{bohg2013data}. 

In recent years, some learning-based approaches have adopted a ``grasp refinement'' step motivated by analytical metrics~\citep{wei2022dvgg,liu2021synthesizing,liu2019generating}. Nevertheless, these metrics tend to be formulated as “relaxed” optimizations to keep the problem tractable. Instead of enforcing the constraints directly, the violation of each constraint is cast as a scalar loss and summed together. This results in an unconstrained optimization which is significantly easier to solve. However, there is no guarantee that the optimization output will indeed satisfy the constraints that motivated the loss design.

\subsection{Simple Grasping}
Due to the geometry of parallel jaw grippers, simple grasp planning can be reduced to computing a 6-DoF gripper pose in space. As such, analytical simple grasp planners may directly reason about the object geometry (e.g.,~\cite{berenson2007grasp}) or rank grasps using grasp quality metrics~\cite{roa2015grasp}. Recently, learning-based approaches that predict wrist poses have gained significant traction (e.g.,~\cite{saxena2006robotic, mousavian20196,mahler2017dex}). We refer the readers to~\citep{du2021vision, roa2015grasp, bohg2013data, caldera2018review, kleeberger2020survey} for a thorough review. These approaches seldom scale directly to dexterous grasp planning, which is a significantly more complex problem.

\subsection{Bilevel Optimization (BO) for Planning}
While BO theory is established in literature \citep{sinha2017review}, application of BO on motion planning is relatively new. BO has been applied to continuous systems~\citep{zimmermann2020multi}, robotic locomotion~\citep{zhu2021contact, landry2019bilevel}, simple pushing and pivoting~\citep{shirai2022robust}, collaborative object manipulation~\citep{stouraitis2020online}, and task and motion planning~\citep{zhao2021sydebo}. To our best knowledge, this work is the first application of BO on dexterous robotic manipulation.
%===============================================================================
\section{Method}

\begin{figure}[!htb]
    \centering
    \includegraphics[width=5.5in]{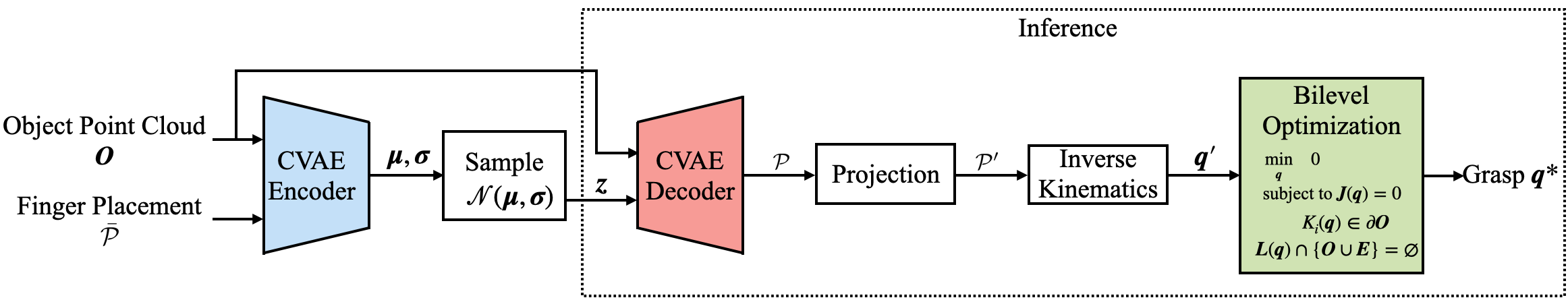}
    \caption{Overview of our method. We train a CVAE that predicts finger placements $\mathcal{P}\in\mathbb{R}^{3\times 3}$ given an object point cloud  $\bm{O}$. At inference time, we first obtain a finger placement prediction $\mathcal{P}$, which is not guaranteed to be physically feasible. Next, we compute a grasp configuration initial guess $\bm{q}'\in\mathbb{R}^{22}$ from $\mathcal{P}$. Finally, we apply BO to compute a physically feasible grasp $\bm{q}^*\in\mathbb{R}^{22}$.}
    \label{fig:overview}
\end{figure}

Our method consists of a learned model that predicts a plausible finger placement $\mathcal{P}\in\mathbb{R}^{3\times 3}$, and an analytical process that computes a \textit{physically feasible dexterous grasp} $\bm{q}^*\in\mathbb{R}^{22}$ (Definition \ref{def:physically_feasible}) guided by $\mathcal{P}$. At inference time, the pipeline takes in an object point cloud and outputs a fully specified grasp configuration $\bm{q}^*$. Figure \ref{fig:overview} gives an overview of our method. We assume the object is grasped with exactly 3 fingers. This section first formally specify the modeled physical constraints, then discuss each of the components in the pipeline. Our implementation is available at \url{github.com/Stanford-TML/dex_grasp}.

\begin{definition}[Physically Feasible Dexterous Grasp]
\label{def:physically_feasible}
    Given an object $\bm{O}$ in environment $\bm{E}$, a grasp configuration $\bm{q}$, and a desired finger placement $\mathcal{P} = (\bm{p}_1, \bm{p}_2, \bm{p}_3)\in\mathbb{R}^{3\times 3}$, we consider $(\bm{O}, \bm{E}, \bm{q}, \mathcal{P})$ to be \textit{physically feasible} if they satisfy following dynamic and kinematic constraints. In our setup, $\bm{q}\in\mathbb{R}^{22}$ and $\bm{E}$ is the fixed tabletop (see Section \ref{sec:hardware_setup}).
\end{definition}

\subsection{Dynamic Constraints: Wrench Closure and Friction Cone} 
For a static dexterous grasp, the dynamic constraints include the wrench closure constraint and the friction cone constraint. The wrench closure constraint requires the sum of the external wrench from all contact points to be zero: 
    $\sum_{i=1}^{3} \bm{f}_i = \bm{0}$ and $\sum_{i=1}^{3} \bm{p}_i \times \bm{f}_i = \bm{0}$.
$\bm{f}_i\in\mathbb{R}^{3}$ is the unknown contact force applied at position $\bm{p}_i\in\mathbb{R}^{3}$ from the dexterous hand to the object.

Given the static friction coefficient $\mu$ and the outward-pointing surface normal $\hat{\bm{n}}_i$, the friction cone constraint requires the normal force to be nonnegative and the contact force to be within the friction cone.
Using a polyhedral cone approximation (\cite{stewart1996implicit}) with an orthogonal basis $\hat{\bm{t}}_{i,j}\perp \hat{\bm{n}}_i, \forall j\in \{1,2\}$, we arrive at the following approximated friction cone constraints:
\begin{equation}
\label{eqn:friction_cone_approx}
    0 \leq -\bm{f}_{i}\cdot \hat{\bm{n}}_i \;\;\; \mathrm{and}\;\;\; |\bm{f}_{i}\cdot\hat{\bm{t}}_{i,j}| \leq -\mu \bm{f}_{i}\cdot \hat{\bm{n}}_i, \forall i\in \{1,2,3\}, \forall j\in \{1,2\}.
\end{equation}

% \begin{equation}
% \label{eqn:friction_cone}
%     0 \leq -\bm{f}_i \cdot \hat{\bm{n}}_i \;\;\; \mathrm{and}\;\;\;\|\bm{f}_i-(\bm{f}_i \cdot \hat{\bm{n}}_i) \hat{\bm{n}}_i\|_2 \leq -\mu \bm{f}_i \cdot \hat{\bm{n}}_i.
% \end{equation}

% The friction force direction may be approximated with a polyhedral cone spanned by a set of unit vectors perpendicular to $\hat{\bm{n}}_i$: $\left\{\hat{\bm{t}}_{i,j}\mid\hat{\bm{t}}_{i,j} \perp \hat{\bm{n}}_i, j\in \{1,2,\dots\} \right\}$ (\cite{stewart1996implicit}). In this work, we chose an arbitrary orthogonal basis on the tangent plane of the contact point $(\hat{\bm{t}}_{i,1}, \hat{\bm{t}}_{i,2})$, which yields the following approximated friction cone constraints
% \begin{equation}
% \label{eqn:friction_cone_approx}
%     0 \leq -\bm{f}_{i}\cdot \hat{\bm{n}}_i \;\;\; \mathrm{and}\;\;\; |\bm{f}_{i}\cdot\hat{\bm{t}}_{i,j}| \leq -\mu \bm{f}_{i}\cdot \hat{\bm{n}}_i, \forall j\in \{1,2\}.
% \end{equation}

A grasp is \textit{dynamically feasible} if it satisfies both wrench closure and friction cone constraints. Leveraging the polyhedral approximation, dynamic feasibility can be cast as a quadratic program (QP):
\begin{align}
\begin{split}
\label{eqn:dyn_feasible_prog}
    \min_{\bm{f}_1, \bm{f}_2, \bm{f}_3\;\;\;}& \|\sum_{i=1}^{3}\bm{f}_i\|^2_2+ \|\sum_{i=1}^{3}\bm{p}_i \times \bm{f}_i\|^2_2
    \;\;\; \textrm{ subject to} \;\;
    0<{f}_{min} \leq -\bm{f}_{i}\cdot \hat{\bm{n}}_i \textrm{ and~\eqref{eqn:friction_cone_approx}}.    
\end{split}
\end{align}
The optimization in Equation \eqref{eqn:dyn_feasible_prog} has objective value $0$ if and only if the grasp is dynamically feasible.
Note that we set an arbitrary lower bound ${f}_{min}\in \mathbb{R}^{+}$ on the normal force to avoid the trivial solution of $\bm{f}_i=\bm{0}$. The exact value of ${f}_{min}$ is irrelevant because changing ${f}_{min}$ represents scaling the optimal contact forces, which does not affect membership in the friction cone.

\subsection{Kinematic Constraints: Reachability and Collision}
The kinematic constraints include reachability constraints and collision constraints. Reachability constraints enforce the kinematic ability of the robot hand to reach the contact point $\bm{p}_i$ on the object surface $\partial\bm{O}$, i.e. $\exists \bm{q}: K_i(\bm{q})= \bm{p}_i\in \partial\bm{O}, \;\forall i\in\{1, 2, 3\}$. Here $K_i: \mathbb{R}^{22} \mapsto \mathbb{R}^3$ is the forward kinematics function that maps the hand state $\bm{q}$ to the position of the $i$-th fingertip. Assuming only fingertip contacts, collision constraints ensure that no robot link $\bm{L}(\bm{q})$ is in collision with the object $\bm{O}$ except at the fingertip or with the environment $\bm{E}$, i.e.
$\bm{L}(\bm{q}) \cap \left\{\bm{O} \cup \bm{E}\right\} = \emptyset.$ Both reachability and collision constraints are nonconvex constraints. %Our implementation handles these constraints using the nonlinear optimizer SNOPT \cite{gill2005snopt} and software provided by Drake \citep{drake}.

\subsection{Learning to Predict Finger Placement from a Physically Feasible Grasp Dataset}
We leverage a generative model to sample potential finger placements $\mathcal{P}$ given an arbitrary object observed as a point cloud $\hat{\bm{O}}$. An immediate challenge faced by the learning approach is the lack of large-scale datasets for dexterous robotic hands. While datasets for real human hands do exist, retargeting human grasping configurations to a robot hand with different kinematic structures and joint limits presents numerous challenges. We opt to synthesize a large-scale dataset of physically feasible grasping configurations for the Allegro robot hand used in this paper.% based on a subset of YCB objects~\citep{calli2017yale}. %$Our dataset is publicly available and can facilitate other research efforts in the field of dexterous manipulation.

\subsubsection{Creating a Physically Feasible Dexterous Grasp Dataset}
\label{subsec:dataset_generation}
We generate physically feasible grasps of six YCB objects \cite{calli2017yale} in simulation. The chosen objects are ``cracker box,'' ``sugar box,'' ``tomato soup can,'' ``mustard bottle,'' ``gelatin box,'' and ``potted meat can.'' We consider a realistic scenario where the object is placed on a flat surface instead of free floating. As such, we need to consider different object rest poses of on the surface in addition to finger placements. In summary, we first generate random object rest poses and enumerate all possible finger placements $\bar{\mathcal{P}}$ using the corresponding simulated object point cloud $\hat{\bm{O}}$. If a kinematically and dynamically feasible grasp $\bm{q}$ can be found for $\bar{\mathcal{P}}$, $(\hat{\bm{O}}, \bm{q}, \bar{\mathcal{P}})$ is added to the dataset $\mathbb{P}$. The detailed procedure is described in the supplementary material.

\subsubsection{Training a Conditional Variational Autoencoder (CVAE)}
To compute diverse grasps for arbitrary objects directly from point cloud observations, we predict fingertip contact points on the object surface with a conditional variational autoencoder (CVAE)~\cite{kingma2013auto}. Our model is adapted from~\citep{mousavian20196} and consists of an encoder $E$ and a decoder $D$ that are based on the PointNet++ architecture~\cite{qi2017pointnet++}. The model seeks to maximize the likelihood of producing a set of contact points $\mathcal{P}$ deemed feasible in Section \ref{subsec:dataset_generation} given the point cloud $\hat{\bm{O}}$.

The encoder $E(\bm{z} \mid \bar{\mathcal{P}}, \hat{\bm{O}})$ maps a grasp $\bar{\mathcal{P}}$ and a point cloud $\hat{\bm{O}}$ to the latent space. We assume the latent variable has a normal distribution $\mathcal{N}(\bm{0},\mathcal{I})$. Meanwhile, given a latent sample $\bm{z}\sim E$, the decoder attempts to reconstruct the finger placement $\mathcal{P}$. We seek to minimize the element-wise $L^1$-norm reconstruction loss $L_{rec}(\mathcal{P}, \bar{\mathcal{P}}) \triangleq \|\mathcal{P}-\bar{\mathcal{P}}\|_1$ for a feasible grasp from the dataset $\bar{\mathcal{P}}\in\mathbb{P}$. Additionally, a KL divergence loss $\mathcal{D}_{KL}$ is applied on the latent distribution $E(\cdot\mid\cdot)$ to ensure a normally distributed latent variable. The complete loss function of the network is defined as
\begin{equation}
    \label{eqn:loss_fn}
    L \triangleq \sum_{\bm{z}\sim E, \bar{\mathcal{P}}\in\mathbb{P}} L_{rec}(\mathcal{P}, \bar{\mathcal{P}}) + \alpha \mathcal{D}_{KL}\left(E(\bm{z} \mid \bar{\mathcal{P}}, \hat{\bm{O}}) \; \middle\vert \middle\vert \; \mathcal{N}(\bm{0}, \mathcal{I})\right).
\end{equation}
At inference time, $E$ is removed and a latent sample $\bm{z}$ is drawn from $\mathcal{N}(\bm{0}, \mathcal{I})$. This is passed into the decoder $D(\mathcal{P}|\hat{\bm{O}}, \bm{z})$ along with the point cloud $\hat{\bm{O}}$ to produce the grasp point prediction $\mathcal{P}$.

\subsection{Computing Grasps with Physical Feasibility Guarantees using Bilevel Optimization (BO)}
\label{sec:bo_main}
While the grasps in the training dataset are physically feasible by construction, there is no guarantee that the CVAE output, $\mathcal{P}$, is physically feasible. Additionally, $\mathcal{P}$ only specifies the finger placement instead of the full hand configuration. %However, projecting $\mathcal{P}$ onto the manifold of physically feasible grasps is challenging as the constraints are highly nonconvex.
We propose a BO to obtain a physically feasible grasping pose $\bm{q}$ given $\mathcal{P}$. To seed the BO with $\mathcal{P}$, we first obtain an Euclidean projection of $\mathcal{P}$ onto $\partial\bm{O}$, denoted as
$
    \mathcal{P}' \triangleq \left\{\bm{p}'_i\; \middle| \; \bm{p}'_i = \argmin_{\bm{p}_o\in\partial\bm{O}}{\|\bm{p}_o-\bm{p}_i\|_2}, \; i\in\{1,2,3\}\right\}.
$
Next, we solve an inverse kinematics (IK) problem for a grasp configuration finger placement problem specified by $\mathcal{P}'$ up to a numerical tolerance $\epsilon$, i.e.
$
\textrm{find } \bm{q}': \|K_i(\bm{q}') - \bm{p}'_i\|_2\leq \epsilon, \; K_i(\bm{q}')\in \partial\bm{O}, \;\forall i\in\{1, 2, 3\}.
$ $K_i:\mathbb{R}^{22}\mapsto\mathbb{R}^3$ is the forward kinematics function of the $i$th finger. $K_i(\bm{q}')\in \partial\bm{O}$ is implemented as a constraint on the finger-object signed distance $D(\bm{p},\bm{O})\in\left[d_{min}, d_{max}\right]$. The IK solution $\bm{q}'$ serves as the initial guess for the BO. 

\subsubsection{Formulating the Bilevel Optimization Grasp Refinement}
\label{sec:method:bilevel}
% To project the 
Accounting for the kinematic and dynamic constraints, the na\"ive formulation of the grasp optimization problem is given in \eqref{eqn:single-level}.
\begin{align}
\begin{split}
\label{eqn:single-level}
    \min_{\bm{q},\bm{f}_1, \bm{f}_2, \bm{f}_3}\;\;\;& 0 \\
    \textrm{subject to}\;\;\;&    K_i(\bm{q}) \in \partial{\bm{O}}, \;
    \bm{L}(\bm{q}) \cap\left\{\bm{O}\cup
    \bm{E} \right\}= \emptyset,\; \|\sum_{i=1}^{3}\bm{f}_i\|_2 = 0,\; \|\sum_{i=1}^{3}K_i(\bm{q}) \times \bm{f}_i\|_2 = 0,\\
    & f_{min} \leq -\bm{f}_{i}\cdot \hat{\bm{n}}_i, \;\;  |\bm{f}_{i}\cdot\hat{\bm{t}}_{i,j}| \leq -\mu \bm{f}_{i}\cdot \hat{\bm{n}}_i, \;\forall i\in \{1,2,3\}, \forall j\in \{1,2\}.
\end{split}
\end{align}
Equation~\eqref{eqn:single-level} does not solve well in practice due to its complexity and nonconvexity. Additionally, while $\bm{q}'$ may serve as an initial guess for $\bm{q}$, the initial guess for $\bm{f}$ is not obvious. Naively relaxing Equation~\eqref{eqn:single-level}, e.g., replacing the objective with
$
    \min_{\bm{q},\bm{f}_1, \bm{f}_2, \bm{f}_3}\; \|\sum_{i=1}^{3}\bm{f}_i\|^2_2+ \|\sum_{i=1}^{3}K_i(\bm{q}) \times \bm{f}_i\|^2_2
$,
may produce suboptimal $\bm{f}$ and, consequently, incorrect conclusion on grasp's dynamic feasibility.
%the guarantee the contact forces are solved to optimality for determining force closure possibility of a chosen $\bm{q}$.

We propose leveraging \textit{bilevel optimization} \citep{sinha2017review,landry2019bilevel} to offload the dynamic feasibility computation from the main optimization. Define $J(\bm{q})$ as the minimum objective value of the dynamic constraint QP in Equation~\eqref{eqn:dyn_feasible_prog} with $p_i=K_i(\bm{q})$:
\begin{equation}
\label{eqn:bilevel_lower}
        J(\bm{q}) \triangleq\min_{\bm{f}_1, \bm{f}_2, \bm{f}_3\;\;\;} \|\sum_{i=1}^{3}\bm{f}_i\|^2_2+ \|\sum_{i=1}^{3}\bm{p}_i \times \bm{f}_i\|^2_2
    \;\;\; \textrm{ subject to} \;\;
    0<{f}_{min} \leq -\bm{f}_{i}\cdot \hat{\bm{n}}_i \textrm{ and~\eqref{eqn:friction_cone_approx}}.
\end{equation}
% \begin{align}
% \label{eqn:bilevel_lower}
% \begin{split}
%     J(\bm{q}) \triangleq \min_{\bm{f}_1, \bm{f}_2, \bm{f}_3}\;\;\;& \|\sum_{i=1}^{3}\bm{f}_i\|^2_2+ \|\sum_{i=1}^{3}K_i(\bm{q}) \times \bm{f}_i\|^2_2\\
%     \textrm{subject to}\;\;\;& 
%     f_{min} \leq -\bm{f}_{i}\cdot \hat{\bm{n}}_i,\;\; %\;\forall i\in \{1,2,3\} \
%     |\bm{f}_{i}\cdot\hat{\bm{t}}_{i,j}| \leq -\mu \bm{f}_{i}\cdot \hat{\bm{n}}_i, \;\forall i\in \{1,2,3\}, \forall j\in \{1,2\}.
% \end{split}
% \end{align}
We observe that the force closure constraint $J(\bm{q})=0$ can be abstracted away to form a ``lower-level'' QP. $J(\bm{q})=0$ can be solved to optimality with exiting QP solvers without reliance on a good initial guess for $\bm{f}_i$.

Applying this abstraction to Equation~\eqref{eqn:single-level} yields the upper-level problem:
\begin{align}
\begin{split}
\label{eqn:bilevel_upper}
    \min_{\bm{q}}\;\;\; 0 \;\;\;
    \textrm{subject to}\;\;\;
    J(\bm{q}) = 0, \
    K_i(\bm{q}) \in \partial{\bm{O}}, \
    \bm{L}(\bm{q}) \cap \left\{\bm{O}\cup\bm{E}\right\} = \emptyset .
\end{split}
\end{align}
Equation~\eqref{eqn:bilevel_upper} is a bilevel optimization as $J(\bm{q})=0$ is a constraint on the minimizer of another optimization problem (Equation~\eqref{eqn:bilevel_upper}). While Equation~\eqref{eqn:bilevel_upper} still defines a nonconvex optimization, the choice for $\bm{f}_i$ has been abstracted away entirely to the lower level QP solver. By construction, a valid solution to Equation~\eqref{eqn:bilevel_upper}, denoted as $\bm{q}^*$, defines a physically feasible grasp.

% We denote the solution to Equation~\eqref{eqn:bilevel_upper} as $\bm{q}^*$. Equation~\eqref{eqn:bilevel_lower} is a constrained QP, it can be solved to optimality numerically with standard convex optimizers. This implies Equation~\eqref{eqn:dyn_feasible_prog} is evaluated exactly with no conservatism in our proposed approach. Moreover, there is no need to provide a good initialization to $\bm{f}_i$.

\subsubsection{Solving the Bilevel Optimization Grasp Refinement}
\label{sec:method:bilevel_solver}
In practice, to solve Equation~\eqref{eqn:bilevel_upper} with a nonconvex optimizer (e.g., SNOPT~\citep{gill2005snopt}), one needs to obtain the gradient of each of the constraints with respect to $\bm{q}$. The gradients of the kinematic constraints can be obtained from simulators with automatic differentiation capabilities (e.g., Drake~\citep{drake}). The gradient of $J(\bm{q})$, $\nabla_{\bm{q}}J(\bm{q})$, can obtained using differentiable optimization \cite{amos2017optnet}. At a high level, differentiable optimization computes $\nabla_{\bm{q}}J(\bm{q})$ by taking the matrix differentials of the KKT conditions of
the optimization problem at its solution. 

We emphasize that the bilevel optimization in Equation~\eqref{eqn:bilevel_upper} is still a highly nonconvex optimization, and the optimizer will likely return a locally optimal $\bm{q}^*$ in the vicinity of $\bm{q}'$. Intuitively, this could be interpreted as optimizing within the grasp family of the CVAE prediction (e.g., ``left side grasp'' or ``top-down grasp''), which implies the necessity of CVAE in the process. This is supported empirically by our ablation studies in Section \ref{sec:evaluation}. In practice, the bilevel optimization provides a certification of physical feasibility for the CVAE prediction. A poor $\bm{q}'$ choice will likely results in the nonconvex optimizer returning infeasibility. This serves as the condition to reject $\mathcal{P}$ and generate a different CVAE grasp prediction with another latent variable sample.

\section{Experiments}
\label{sec:evaluation}
\begin{figure}[h]
    \centering
     \begin{subfigure}[t]{0.19\textwidth}
         % \centering
         \includegraphics[width=\textwidth]{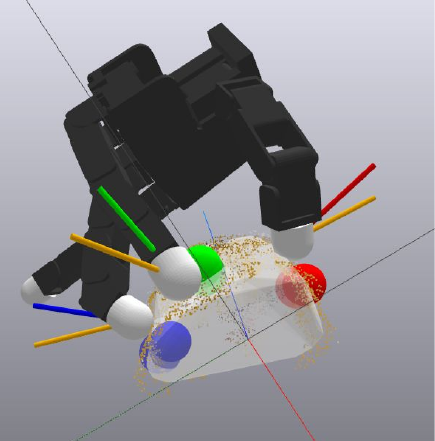}
         \caption{Before BO ($\bm{q}'$).}
         \label{fig:bilevel:before}
     \end{subfigure}
     \hfill
     \begin{subfigure}[t]{0.19\textwidth}
         % \centering
         \includegraphics[width=\textwidth]{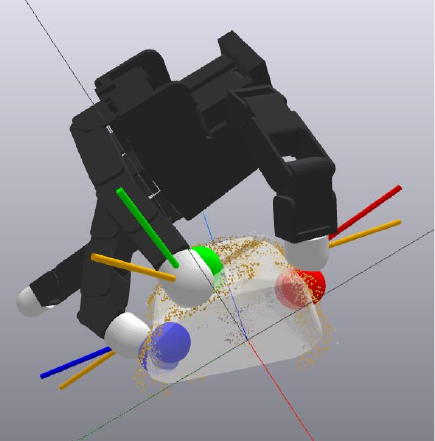}
         \caption{After BO ($\bm{q}^*$).}
         \label{fig:bilevel:after}
     \end{subfigure}
    \begin{subfigure}[t]{0.29\textwidth}
        \includegraphics[width=\textwidth]{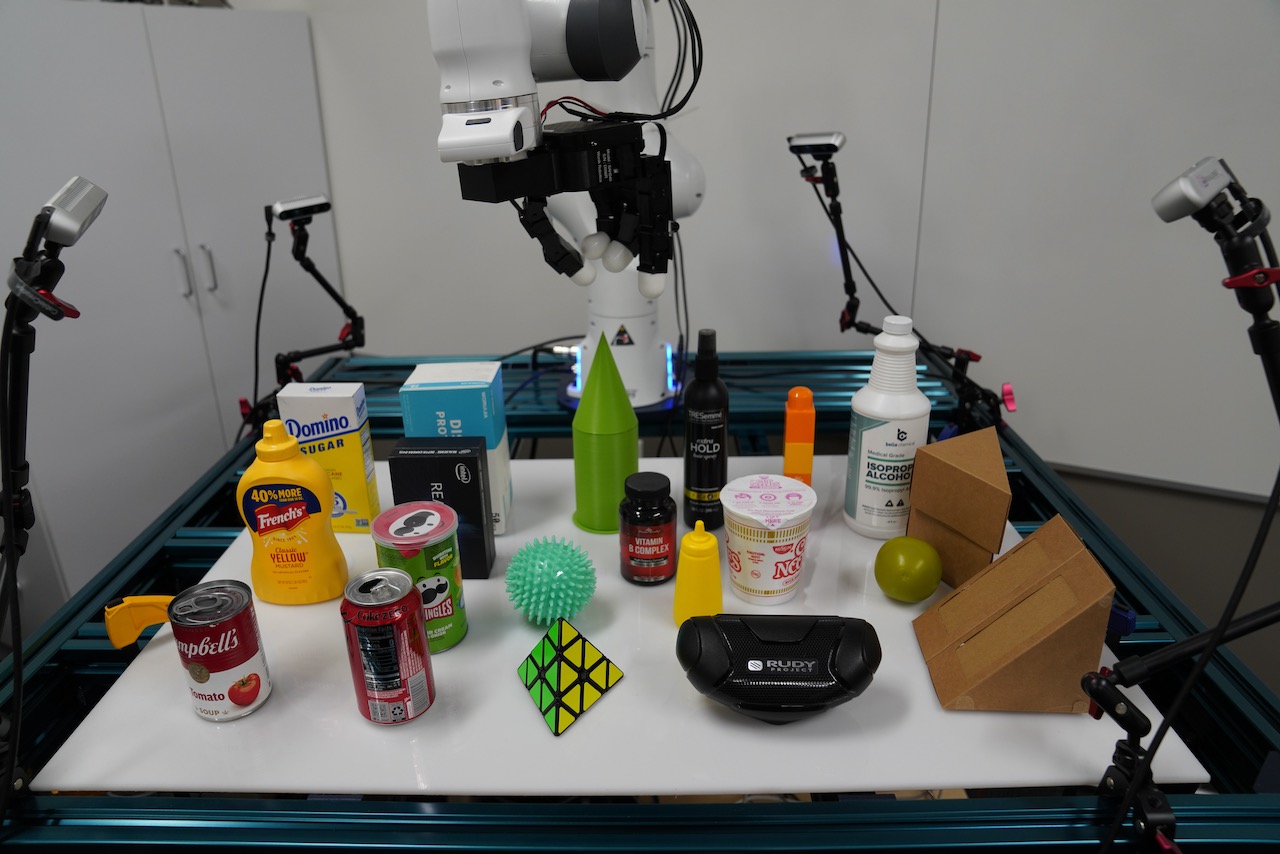}
        \caption{Hardware setup.}
        \label{fig:hardware_setup}     
    \end{subfigure}
    \begin{subfigure}[t]{0.31\textwidth}
        \includegraphics[width=\textwidth]{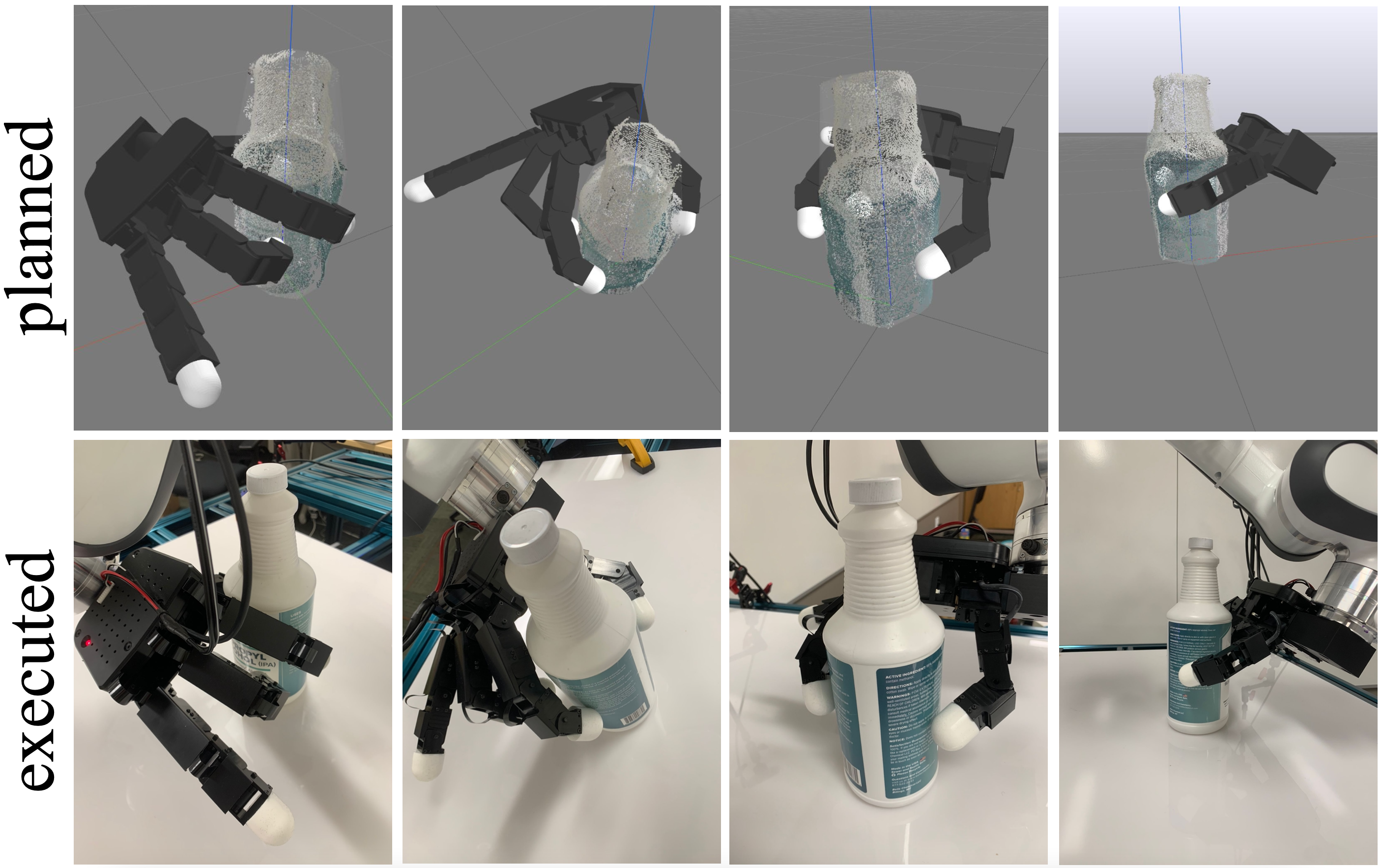}
        \caption{Planned and executed grasps. 
        }
        \label{fig:sim_hw_correspondence}
    \end{subfigure}

\caption{\textbf{\ref{fig:bilevel:before} and \ref{fig:bilevel:after}: Before and after BO}, on a bottom-up view of a mustard bottle grasp.
    Using the color encoding of red: thumb, green: index, and blue: middle, we show $\mathcal{P}$ (solid spheres), $\mathcal{P}'$ (transparent spheres) and $\hat{\bm{n}}_i$ (colored lines). The direction of the computed contact forces $\bm{f}_i$ are shown at the respective fingertip with yellow lines. The mismatch between $K(\bm{q}')$ and $\mathcal{P}'$ is due to the numerical tolerance $\epsilon$.
    \textbf{\ref{fig:hardware_setup}: Hardware setup.} See Section~\ref{sec:hardware_setup} for more details.
    \textbf{\ref{fig:sim_hw_correspondence}: Grasp sim-to-real.} Correspondence between the planned grasp in simulation and execution on hardware. 
    }
    \label{fig:bilevel_sim}
\end{figure}

Our trained CVAE achieved a test reconstruction error of $0.5$cm. More training details are available in the supplementary material. The effects of applying BO is shown in Figures \ref{fig:bilevel:before} and \ref{fig:bilevel:after}. The finger placement prior to BO cannot form force closure as $\hat{\bm{n}}_i\cdot \hat{\bm{n}}_j > 0, \forall i,j\in\{1,2,3\}$. BO shifted the thumb and middle fingers to an antipodal configuration, which allows for force closure. This validates our approach's ability to make an initially infeasible grasp prediction physically feasible. The rest of this section focuses on hardware experiments and evaluations.

\subsection{Hardware Setup}
\label{sec:hardware_setup}
The hardware setup and evaluated objects are shown in Figure \ref{fig:hardware_setup}.\\
\textbf{Hardware.} A 16-DoF Allegro v4.0 right hand was used for hardware grasping experiments. The Allegro hand was mounted on a 7-DoF Franka Emika Panda arm to realize the planned wrist pose.
The object point cloud was captured by 4 stationary Intel RealSense D435 depth cameras. The implementation details are available in the supplementary material.\\
\textbf{Evaluated Objects.} We evaluated our method on 20 rigid household objects resting on the table. We categorize our objects into three sets:\\
$\bullet\;\,$\textbf{3 Seen objects.} (Leftmost column) mustard bottle, soup can, and sugar box from the training set.\\
$\bullet$ \textbf{4 Familiar objects.} (Second-from-left column) Unseen objects with geometries similar to training objects. Includes boxes (webcam box, mask box) and cylinders (chip can, soda can).\\
$\bullet$ \textbf{13 Novel objects.} Objects whose geometries are distinct from that of any object in the training set. From left to right, front to back: tetrahedron, massage ball, castle, pill bottle, glasses case, condiment bottle, hairspray bottle, ramen, lego, sandwich box (side), pear, sandwich box (upright), alcohol bottle.\\

\subsection{Experiment Procedure}
At the start of each trial, the object is placed in a specified pose in front of the robot. After observing the point cloud and computing a grasp, the Franka arm first brings the hand to the desired wrist pose using an \textit{ad hoc} trajectory planner. The finger joint angles $\bm{\theta}^*\in\mathbb{R}^{16}$ from $\bm{q}^*$ is then executed on the Allegro hand. Contact forces are provided by squeezing the fingertips according to the planned contact forces: $\bm{\theta}^* \gets \bm{\theta}^*+k(\nabla_{\bm{\theta}}\bm{p}_i)^T\bm{f}_i$. Here $\nabla_{\bm{\theta}}\bm{p}_i$ is the Jacobian of the fingertip locations with respect to $\bm{\theta}$, and $k$ is a fixed ``stiffness'' constant as motivated by impedance control. The Franka then attempts to lift the object to a fixed height approximately $43$cm above the table. A trial is considered successful if the object is lifted and all three fingers remain in contact with the object. 
3 repeats are performed for hardcoded policy trials as they are deterministic up to the object point cloud observation. All other trials are repeated 6 times. Our hardware pipeline was implemented with the intent to execute our planned grasp as accurately as possible. Figure~\ref{fig:sim_hw_correspondence} illustrates the simulation-hardware grasp correspondence.

% \begin{figure}[t]
%     \centering
%     \includegraphics[width=0.3\textwidth]{figures/sim2real.pdf}
%     \caption{Correspondence between the planned grasp in simulation and hardware execution. 
%     }
%     \label{fig:sim_hw_correspondence}
% \end{figure}

\subsection{Baseline and Ablation Studies}

We compare our method against the following approaches. We excluded ``BO only'' in our ablation studies as it does not return a result without CVAE in practice. BO seldom returns a solution if seeded with a kinematically infeasible solution (e.g., open hand).\\
$\bullet$ \textbf{Hardcoded grasp baseline.} Using the set of seen objects, we designed a top-down tripod grasp policy that chooses the wrist pose based on the point cloud and executes a fixed grasp. \\
$\bullet$ \textbf{CVAE-only as an ablation study.} We solved a collision-free inverse kinematics problem that attempts to match the CVAE predicted fingertip positions $\mathcal{P}$. This mimics a ``learning only'' approach.
\\
$\bullet$ \textbf{CVAE-kinematic as an ablation study.} We ablated the ``lower level'' part of the optimization (Equation \eqref{eqn:bilevel_lower}) away and use $\bm{q}'$ directly without bilevel optimization. This ablation considers kinematic constraints but not dynamic constraints.

\subsection{Grasp Trial Results}
\textbf{Our method achieved an overall success rate of 86.7\%} over 120 grasp trials on 20 objects. This is superior to the hardcoded baseline, which achieved an overall success rate of $53.3\%$. On seen objects, our CVAE-only ablation achieved a $38.9\%$ success rate and our CVAE-kinematic ablation achieved an $88.9\%$ success rate. The results are summarized in Table \ref{tab:grasp_statistics}, and the details are available in the supplementary material. A video summary is available at  \url{youtu.be/9DTrImbN99I}. 

\textbf{Our method can grasp challenging novel objects}. 
%For instance, the sandwich box is challenging to parallel jaw grippers as it is triangular. The (rigid) massage ball is challenging to suction-based grippers as the surface is highly irregular. The castle object requires a side grasp due to its cone-shaped top. Slender objects such as lego and hairspray bottle also benefit from a side grasp as there is more room for finger placement along the vertical axis compared to a horizontal cross section. The tetrahedron is difficult for any finger-based gripper to achieve force closure due to its shape, and a large squeezing force is necessary. 
%Our method achieved high grasp success rates on all objects except for the tetrahedron. 
This includes objects that are difficult for parallel grippers and suction cups to grasp. We discuss the challenges of some of our test objects in the supplementary material. The hardcoded baseline nearly always succeeds on objects that allow top-down grasps and are similar in size to the seen object used for tuning. However, the baseline fails on objects that either require a different grasp type or have significantly different size.

\begin{figure}[]
    \centering
    \includegraphics[width=0.85\textwidth]{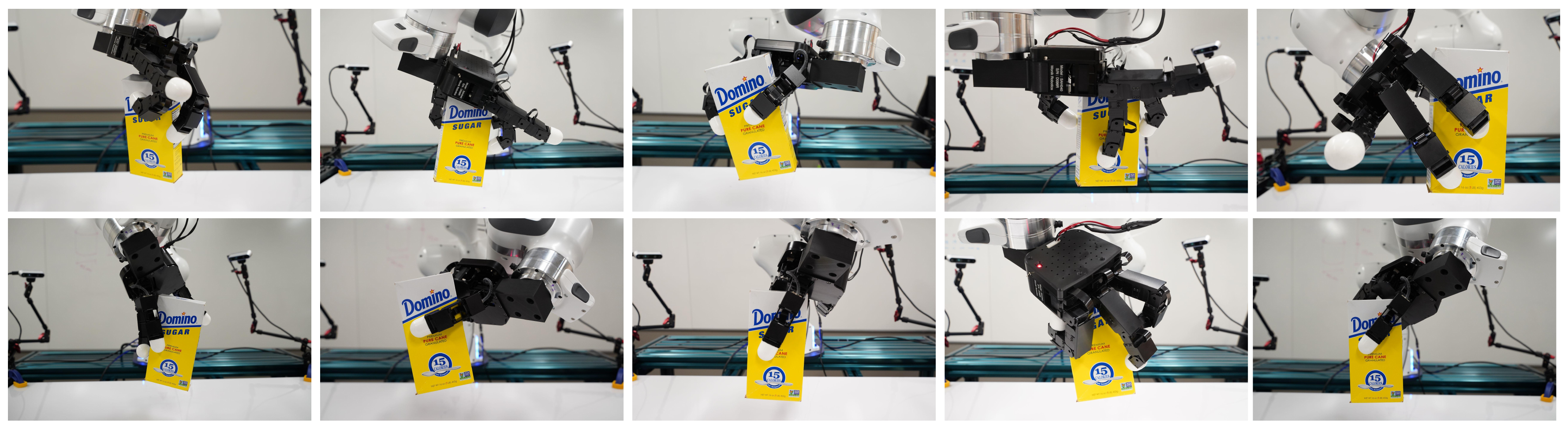}
    \caption{Diverse grasps generated by our method. Each image is a unique grasp generated from different sampled latent variables. All grasps were computed for the same initial object rest pose.
    }
    \label{fig:diverse_grasps}
\end{figure}

\textbf{Our method can generate diverse grasps through sampling different latent $\bm{z}$'s.} Figure~\ref{fig:diverse_grasps} shows 10 distinct successful grasps performed on the upright sugar box.

\textbf{The median time to produce a grasp with our method is 14.4 seconds}. The median repeats for each component of our method when generating a grasp is two CVAE $\bm{z}$ samples, two IK solves for $\bm{q}'$, and two bilevel optimizations for $\bm{q}^*$. Detailed timing and repetition results are available in the supplementary material.

\begin{table}[h]
\centering\scriptsize
\caption{Object grasping statistics. Our method achieved superior grasp success rate compared to the ablations and the hardcoded baseline.}
\begin{tabular}{|c|c|c|c|c|c|c|c|c|}
\hline
 Category (object count) & \multicolumn{2}{|c|}{Seen (3)} & \multicolumn{2}{|c|}{Familiar (4)} & \multicolumn{2}{|c|}{Novel (13)} & \multicolumn{2}{|c|}{Overall (20)} \\ \hline
Ours & 17/18 & 94.4\% & 23/24 & 95.8\% & 64/78 & 82.1\% &104/120 & 86.7\%\\ \hline
CVAE only & 7/18 & 38.9\% & - & - & - & - & - & - \\ \hline
CVAE+kinematics & 16/18 & 88.9\% & - & - & - & - & - & - \\ \hline
Hardcoded & 8/9 & 88.9\% & 9/12 & 75.0\% & 15/39 & 38.5\% & 32/60 & 53.3\%\\ \hline
\end{tabular}
\label{tab:grasp_statistics}
\end{table}

\subsection{Quantitative evaluation of physical constraint enforcement}
\label{sec:feasibility}
Table~\ref{tab:constr_stats} summarizes the quantitative evaluation of physical constraints.\\ 
\textbf{Kinematic constraints.} The maximum finger-object distance constraint violation in a grasp planned by our method is 0.08cm, which is negligible in practice as it is smaller than other error sources such as camera observation error. The CVAE-kinematics ablation, which only considers kinematic constraints, achieved comparable results. The CVAE-only ablation has a maximum violation of 2.032cm, confirming that while CVAE alone can produce qualitatively correct grasp, it cannot enforce kinematic constraints precisely.\\
\textbf{Dynamic constraints}. To evaluate dynamic constraint satisfaction, we chose ``force (torque) ratio'' as our metric, i.e.
%the ratio between the applied external force(torque) versus the average magnitude of the applied contact force(torque) 
$
    \frac{\|\sum_{i=1,2,3}\bm{f}_i\|_2}{\frac{1}{3}\sum_{i=1,2,3}\|\bm{f}_i\|_2}\times 100\%,
$
and
$
    \frac{\|\sum_{i=1,2,3}\bm{p}_i \times \bm{f}_i\|_2}{\frac{1}{3}\sum_{i=1,2,3}\|\bm{p}_i \times \bm{f}_i\|_2}\times 100\%.
$
If wrench closure is achieved, both ratios should be zero. All ratios are reported as \textit{median, (25th percentile, 75th percentile)}. We computed these ratios for grasps planned with our method and CVAE-kinematics ablations. These ratios are not computed for CVAE-only because there are often fewer than 3 finger-object contacts. %The results are summarized in Table~\ref{tab:constr_stats}.
Our method achieved a median of $\leq 0.01\%$ on both ratios, showing that BO is effective in enforcing wrench closure. We note that while the CVAE+kinematics ablation does not explicitly consider external wrench, it still produced many grasps that achieve wrench closure by coincidence. This reflects CVAE’s ability to produce \textit{qualitatively} correct grasps.

\begin{table}[]
\caption{Evaluation of physical constraint on various grasps. A physically feasible grasp should satisfy $D(\bm{p},\bm{O})\in\left[d_{min}, d_{max}\right] = \left[-0.68, -0.32\right]$ and zero force and torque ratios.}
\centering\scriptsize
\begin{tabular}{|c|c|c|c|c|c|c|}
\hline
&Median $D$ & Min $D$ & Max $D$ & Max violation & Force ratio & Torque ratio \\ \hline
Ours (all objects) & -0.32 & -0.68 & -0.24 & 0.08&(0.00, 0.02) & 0.01 (0.00, 2.68) \\ \hline
CVAE-kinematics (seen objects) & -0.32 & -0.64 & -0.32 & 0.04 &0.00 (0.00, 0.12) & 0.01 (0.00, 19.54) \\ \hline
CVAE-only (seen objects) & 0.33 & -1.10 & 1.71 & 2.03 & - & -\\
\hline

\end{tabular}
\label{tab:constr_stats}
\end{table}

\begin{table}[]
\caption{Correlation between dynamic feasibility and hardware success. Only the grasps plans with wrench closure resulted in successful hardware execution.}
\centering\scriptsize
\begin{tabular}{|c|c|c|c|c|}
\hline
& \multicolumn{2}{c}{Successful} & \multicolumn{2}{|c|}{Failed} \\
\hline
&Force ratio & Torque ratio &Force ratio & Torque ratio \\
\hline
BO rejected (12 of 12 failed)&-& -& 61.81 (25.51, 91.65)
& 48.53 (39.89, 78.68) \\ \hline
CVAE-kinematics (7 of 12 failed)& 0.00 (0.00, 0.00) & 0.00 (0.00, 0.01) & 50.71 (19.13, 128.85) & 36.75 (30.48, 131.82)\\ \hline
\end{tabular}
\label{tab:dynamic_feasibility_success}
\end{table}

To show that physical feasibility is a necessary condition for a successful grasp, we examined the CVAE+kinematic trials, which may not satisfy dynamic constraints, on “ramen” and “mustard bottle.” We also executed grasp plans that are reported to be infeasible by BO. Table~\ref{tab:dynamic_feasibility_success} summarizes the results. There is a strong correlation between hardware success and wrench closure. Moreover, all grasps reported to be infeasible by BO failed on hardware. This confirms that grasp refinement derived from rigorous physics-inspired metrics can significantly improve the final grasp plan quality.

% We report statistics on the \textit{maximum of the force and torque ratio} of these grasps, i.e. $
%     \max\left\{ \frac{\|\sum_{i=1,2,3}\bm{f}_i\|_2}{\frac{1}{3}\sum_{i=1,2,3}\|\bm{f}_i\|_2},    \frac{\|\sum_{i=1,2,3}\bm{p}_i \times \bm{f}_i\|_2}{\frac{1}{3}\sum_{i=1,2,3}\|\bm{p}_i \times \bm{f}_i\|_2}\right\}, 
% $
% in Table~\ref{tab:bilevel_reject}.

% \begin{table}[]
% \caption{\revised{Statistics on the \textit{maximum of the force and torque ratios}, computed for grasps either 1. rejected by bilevel optimization 2. produced by CVAE-kinematics and reported as \textit{median, (25th percentile, 75th percentile)}. Grasps produced by each approach are split based on their hardware execution outcome. None of the bilevel rejected grasps succeeded on hardware, while 7 of 12 grasps planned by CVAE-kinematics succeeded. All successful grasps have low force/torque ratios, while all failed grasps have high force/torque ratios.}}
% \centering\scriptsize
% \begin{tabular}{|c|c|c|}
% \hline
%  & Successful & Failed \\ \hline
% Bilevel rejected (0/12 success) & N/A & 65.90 (42.33, 101.40) \\ \hline
% CVAE-kinematics (7/12 success) & 0.00 (0.00, 0.01) & 50.71 (30.48, 131.82) \\ \hline
% \end{tabular}
% \label{tab:bilevel_reject}
% \end{table}

%===============================================================================

\section{Limitations and Conclusion}
\paragraph{Limitations.} Our method requires an observation of the full object point cloud. As our choice of physical constraint formulation does not include explicit robustness margins, estimation errors from the vision pipeline is a major source of grasp failures. This may be addressed by introducing point cloud completion (e.g.,~\citep{yuan2018pcn, sarmad2019rl, han2017high}) or grasp robustness metrics (e.g.,~\citep{ferrari1992planning}). Additionally, it is currently not possible to specify which type grasps to generate (e.g., ``top grasp'' or ``side grasp'') with our method. Finally, our method currently assumes that the object to grasp is placed on a flat surface and that there are no other objects in the scene.

\paragraph{Conclusion.} This work presents a novel pipeline that combines learning-based grasp generation with bilevel optimization to produce diverse and physically feasible dexterous grasps. Our method achieved a grasp success rate of $86.7\%$ on 20 challenging real-world objects. Ablation studies demonstrated that an integrative approach combining learned models and rigorous physics-inspired metrics can achieve superior grasp output quality. Grasps initially generated by CVAE may not satisfy all physical constraints. However, by incorporating bilevel optimization for grasp refinement and rejection sampling, poor grasp predictions can be removed. This paradigm of combining learning and physics-inspired bilevel optimization may be applied to other robotic manipulation tasks.

\clearpage
% The acknowledgments are automatically included only in the final and preprint versions of the paper.
\acknowledgments{We would like to thank Oussama Khatib, Jeannette Bohg, Dorsa Sadigh, Samuel Clarke, Elena Galbally Herrero, Wesley Guo, and Yanchao Yang for their assistance on setting up the hardware experiments. We would also like to thank Chen Wang and Jiaman Li for advice on training the CVAE model. Our research is supported by NSF-NRI-2024247, NSF-FRR-2153854, Stanford-HAI-203112, and the Facebook Fellowship.}
% \acknowledgments{If a paper is accepted, the final camera-ready version will (and probably should) include acknowledgments. All acknowledgments go at the end of the paper, including thanks to reviewers who gave useful comments, to colleagues who contributed to the ideas, and to funding agencies and corporate sponsors that provided financial support.}

%===============================================================================

% no \bibliographystyle is required, since the corl style is automatically used.
\bibliography{biblio}  % .bib
% 	Citations can be made using either \textbackslash citep\{\} or \textbackslash citet\{\}, depending from the appropriateness. To avoid the citation moving to the next line, it is often a good practice to replace the space before with a tilde (\~{}) character.
% 	Example 1: ``CoRL is the best conference ever~\citep{Gauss1857}.''
% 	Example 2: ``\citet{Gauss1857} proved, both theoretically and numerically, that CoRL is the best conference ever.''

% \clearpage
% \appendix

\end{document}

\maketitle

\section{Method}
\subsection{Polyhedral Approximation of the Friction Cone}
The friction cone can be modeled as in Equation~\eqref{eqn:friction_cone}.
\begin{equation}
\label{eqn:friction_cone}
    0 \leq -\bm{f}_i \cdot \hat{\bm{n}}_i \;\;\; \mathrm{and}\;\;\;\|\bm{f}_i-(\bm{f}_i \cdot \hat{\bm{n}}_i) \hat{\bm{n}}_i\|_2 \leq -\mu \bm{f}_i \cdot \hat{\bm{n}}_i.
\end{equation}

The friction force direction may be approximated with a polyhedral cone spanned by a set of unit vectors perpendicular to $\hat{\bm{n}}_i$: $\left\{\hat{\bm{t}}_{i,j}\mid\hat{\bm{t}}_{i,j} \perp \hat{\bm{n}}_i, j\in \{1,2,\dots\} \right\}$ (\cite{stewart1996implicit}). In this work, we chose an arbitrary orthogonal basis on the tangent plane of the contact point $(\hat{\bm{t}}_{i,1}, \hat{\bm{t}}_{i,2})$, which yields the following approximated friction cone constraints
\begin{equation}
\label{eqn:friction_cone_approx}
    0 \leq -\bm{f}_{i}\cdot \hat{\bm{n}}_i \;\;\; \mathrm{and}\;\;\; |\bm{f}_{i}\cdot\hat{\bm{t}}_{i,j}| \leq -\mu \bm{f}_{i}\cdot \hat{\bm{n}}_i, \forall j\in \{1,2\}.
\end{equation}

\subsection{Enforcing the Finger-Object Distance Constraint}
The finger-object distance constraint 
$K_i(\bm{q}')\in \partial\bm{O}$ is implemented as a constraint on the signed distance $d_{min} \leq D(K_i(\bm{q}'), \bm{O}) \leq d_{max}$, where $D$ is defined as
\begin{equation}
\label{eqn:signed_distance}
     D(\bm{p}, \bm{O}) \triangleq \begin{cases}
    \min_{\bm{p}_o\in\partial\bm{O}}{\|\bm{p}_o-\bm{p}\|_2},& \text{if } \bm{p}\notin \bm{O}\\
    -\min_{\bm{p}_o\in\partial\bm{O}}{\|\bm{p}_o-\bm{p}\|_2},              & \text{otherwise.}
     \end{cases}
\end{equation}
As the fingertips of the allegro hand are compliant, we chose $d_{min}=-0.68$cm and $d_{max}=-0.32$cm based on the thickness of the compliant material. 

\subsection{Creating a Physically Feasible Dexterous Grasping Dataset}
\label{subsec:dataset_generation}
We generate physically feasible grasps of six YCB objects \cite{calli2017yale} in simulation. The chosen objects are ``cracker box,'' ``sugar box,'' ``tomato soup can,'' ``mustard bottle,'' ``gelatin box,'' and ``potted meat can.'' We consider a realistic scenario where the object is placed on a flat surface instead of free floating. As such, we need to consider the variations due to different rest poses of on the surface in addition to finger placements. The following summarizes the steps for data generation:
\begin{enumerate}
\item Simulate dropping the object on a flat surface from different initial poses to find ``rest poses''.
\item For each rest pose, produce 256 candidate contact points distributed across the object with Poisson disk sampling.
\item Remove any point in contact with the surface $\bm{E}$ and store the remaining as a point cloud $\hat{\bm{O}}$.
\item Evaluate Equation \eqref{eqn:dyn_feasible_prog} for each of the $C_{3}^{256}$ 3-point finger placements. Discard if it is not dynamically feasible.
\begin{align}
\begin{split}
\label{eqn:dyn_feasible_prog}
    \min_{\bm{f}_1, \bm{f}_2, \bm{f}_3\;\;\;}& \|\sum_{i=1}^{3}\bm{f}_i\|^2_2+ \|\sum_{i=1}^{3}\bm{p}_i \times \bm{f}_i\|^2_2\\
    \textrm{subject to }\;\;\;& 
    0<{f}_{min} \leq -\bm{f}_{i}\cdot \hat{\bm{n}}_i, \;\;|\bm{f}_{i}\cdot\hat{\bm{t}}_{i,j}| \leq -\mu \bm{f}_{i}\cdot \hat{\bm{n}}_i, \;\forall i\in \{1,2,3\}, \forall j\in \{1,2\}.
\end{split}
\end{align}
\item For each dynamically feasible finger placement, permute the $3!=6$ possible assignments for the thumb, index finger, and middle finger. Check each assignment for kinematic feasibility. If a feasible grasping pose $\bm{q}$ can be found for a finger placement $\bar{\mathcal{P}}$, add $(\hat{\bm{O}}, \bm{q}, \bar{\mathcal{P}})$ to the dataset $\mathbb{P}$.
\item At training time, augment the dataset by randomly translating and varying the yaw angle for each grasp in the dataset.
\end{enumerate}

\section{Software}
\subsection{Implementation Details}
The CVAE is implemented using PyTorch~\cite{paszke2019pytorch} and trained on a Google Cloud virtual machine instance with 16 NVIDIA Tesla A100 GPUs. The selected model took approximately 76 hours to train.

The bilevel optimization pipeline is implemented using a mixture of automatic differentiation tools from Drake~\cite{drake}, OptNet~\cite{amos2017optnet}, and Pytorch~\cite{paszke2019pytorch}.

Our source code is available at \url{github.com/Stanford-TML/dex_grasp}.

\subsection{CVAE Training Process}

\begin{figure}[h]
    \centering
     \begin{subfigure}[b]{0.4\textwidth}
         \centering
         \includegraphics[width=\textwidth]{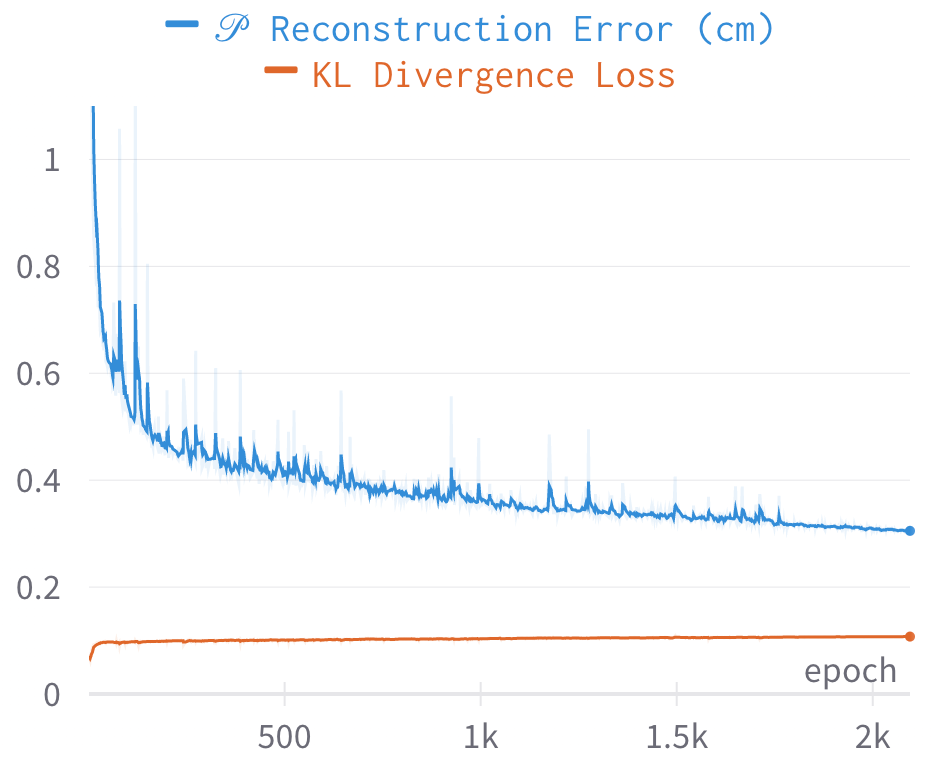}
         \caption{CVAE Training Loss.}
         \label{fig:cvae_training:training_loss}
     \end{subfigure}
     \hfill
     \begin{subfigure}[b]{0.4\textwidth}
         \centering
         \includegraphics[width=\textwidth]{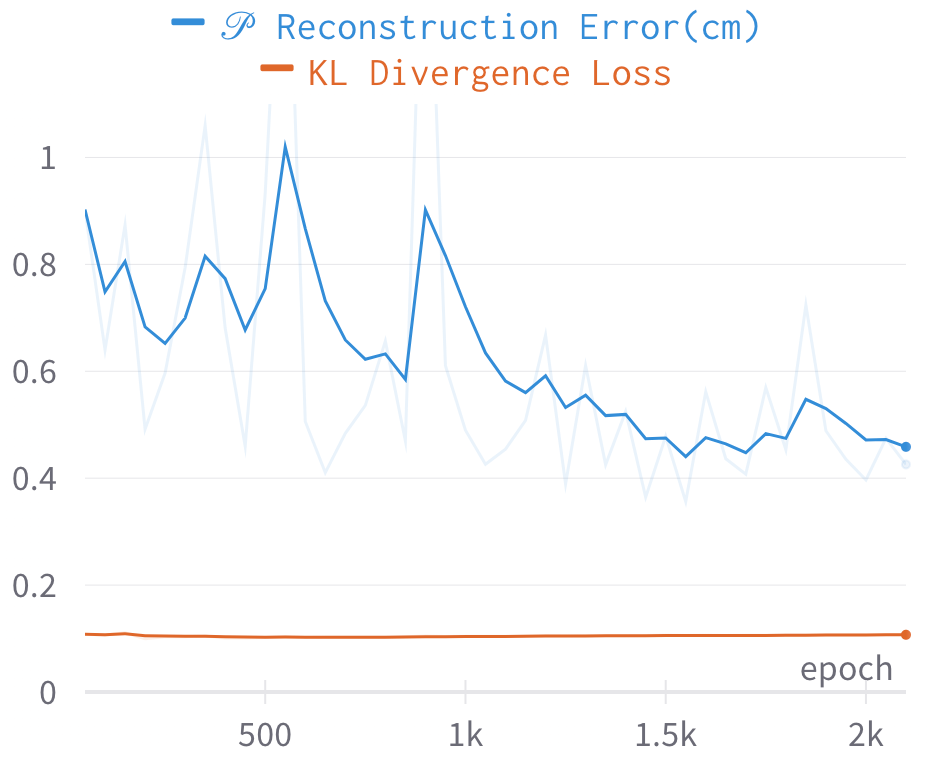}
         \caption{CVAE Testing Loss.}
         \label{fig:cvae_training:testing_loss}
     \end{subfigure}
    \caption{Training and testing loss of the CVAE. The model was able to achieve sub-centimeter reconstruction accuracy of the fingertip placement. Based on the testing results, the model weights at $1550$ training epochs were selected for all subsequent experiments.}
    \label{fig:cvae_training}
\end{figure}

Figure~\ref{fig:cvae_training} shows the training curve. Based on the test losses, we chose the model after 1550 epochs of training for all subsequent experiments.

\section{Hardware}
\subsection{Robot Control} 
The Franka arm was controlled using Polymetis~\cite{Polymetis2021}. The Allegro hand was commanded with an open source driver~\cite{allegro_linux}.

\subsection{Scene observation} 
We collect depth images from the four calibrated cameras to obtain a point cloud of the tabletop scene in the robot base frame.
To obtain a segmented object point cloud, we remove tabletop points (via plane fitting), crop points within a predefined axis-aligned bounding box (roughly corresponding to a small region near the center of the table), and filter the point cloud to remove outliers~\cite{Zhou2018}.
Finally, we extract a mesh from the object point cloud by computing its convex hull.

\subsection{CVAE-based methods} For CVAE-based methods, which require sampling latent variables, we utilize unique random seeds across trials.
Within each trial, we sample latent variables until we encounter the first successful grasp in simulation (kinematic and dynamic constraints satisfied).
This is the grasp that we execute and evaluate on hardware.
Each sampled latent variable attempts six IK initializations (orientations) before the next latent variable is sampled.
% \karen{Need to report how many attempts on average before we find a good grasp.}

% \subsection{Experiment procedure}
% At the start of each trial, the object is placed in one of its specified poses at the center of the table, in front of the robot.
% The arm and hand are initialized to ``home'' poses, in which the wrist is approximately $43$cm above the table.
% To execute a grasp, the hand travels to the planned wrist pose via a hand-designed approach trajectory.
% The planned grasp is executed open loop at $20$ Hz with an \textit{ad hoc} squeezing motion generated by moving the fingers from $\bm{q}_\text{home}$ to $\bm{q}_\text{squeeze} = 1.3(\bm{q}-\bm{q}_\text{home})$, where $\bm{q}$ is computed by our method and $\bm{q}_\text{home}$ is the hardcoded ``home'' pose. 
% \karen{Can we get rid of i?}
% The extrapolation of $\bm{q}$ is needed to provide sufficient contact forces to lift the object. 
% After grasp execution, the hand returns to the home wrist pose while grasping the object.

% \subsection{Evaluation metric} 
% We report the grasp success rate across all trials.
% A trial is considered successful if the object is lifted and all three fingers are in contact with the object.

% We exclude any trials where the robot is unable to match the planned grasp pose due to either (i) hardware failure, (ii) collision of the arm or hand while approaching the planned wrist pose, or (iii) collision of the fingers with the tabletop during grasp execution. 

\subsection{Challenges Stemming from Selected Object Shapes}
The sandwich box is challenging to parallel jaw grippers as it is triangular. The (rigid) massage ball is challenging to suction-based grippers as the surface is highly irregular. The castle object requires a side grasp due to its cone-shaped top. Slender objects such as lego and hairspray bottle also benefit from a side grasp as there is more room for finger placement along the vertical axis compared to a horizontal cross section. The tetrahedron is difficult for any finger-based gripper to achieve force closure due to its shape, and a large squeezing force is necessary. 
%Our method demonstrated successful diverse dexterous grasp generation by achieving high grasping success rates on all objects except for the tetrahedron. While our hardcoded baseline nearly always succeeds on objects that 1. allow top-down grasps and 2. is similar in size to the seen object used for tuning, it fails on objects that either require a different grasp type or have significantly different size.

\subsection{Hardcoded Baseline Design}
The grasp finger joint angles are manually specified. The wrist translation is defined as $x=\frac{1}{N}\sum_{i=1}^N x_i+x_\text{offset}$, $y=\frac{1}{N}\sum_{i=1}^N y_i+y_\text{offset}$, and $z=\max_{i} z_i + z_\text{offset}$, where $(x_\text{offset}, y_\text{offset}, z_\text{offset})$ are hand-tuned offsets and $(x_i, y_i, z_i)$ is the $i$th point in the point cloud containing $N$ points.
The orientation of the wrist is aligned with the short axis obtained from performing principal component analysis on the $x$-$y$ projection of the observed point cloud.
% \item \textbf{Parallel jaw gripper (PJG).} We emulate a PJG with the Allegro hand by using only two of the three fingers from the hardcoded baseline. \mg{Only keep this if we get results for it.}
% Note that we also tested a variant of our proposed method by seeding the bilevel optimization with the ``home'' pose without CVAE prediction. In this case, bilevel optimization seldom returns a solution, so we did not complete the full hardware trial.

\section{Results}
\subsection{Grasp Trial Statistics}
Tables~\ref{tab:seen_obj},~\ref{tab:familiar_obj},~\ref{tab:novel_obj} summarize the detailed success rates on each of the 20 test objects.
\begin{table}[h]
\centering\scriptsize
\caption{Seen objects grasp statistics.}
\begin{tabular}{|c|c|c|c|c|c|}
\hline
 & Sugar box & Soup Can & Mustard Bottle & Overall & \textbf{Success \%} \\ \hline
Ours & 6/6 & 6/6 & 5/6 & 17/18 & \textbf{94.4} \\ \hline
CVAE only & 3/6 & 3/6 & 1/6 & 7/18 & \textbf{38.9} \\ \hline
CVAE+kinematics & 6/6 & 6/6 & 4/6 & 16/18 & \textbf{88.9} \\ \hline
Hardcoded & 3/3 & 2/3 & 3/3 & 8/9 & \textbf{88.9} \\ \hline
\end{tabular}
\label{tab:seen_obj}
\end{table}

\begin{table}[h]
\centering\scriptsize
\caption{Familiar objects grasp statistics.}
\begin{tabular}{|c|c|c|c|c|c|c|}
\hline
 & Webcam & Mask & Chips & Soda & Overall & \textbf{Success \%} \\ \hline
Ours & 6/6 & 6/6 & 5/6 & 6/6 & 23/24 & \textbf{95.8} \\ \hline
Hardcoded & 3/3 & 3/3 & 2/3 & 1/3 & 9/12 & \textbf{75.0} \\ \hline
\end{tabular}
\label{tab:familiar_obj}
\end{table}

\begin{table}[h]
\centering\scriptsize
\caption{Novel objects grasp statistics.}
\begin{tabular}{|c|c|c|c|c|c|}
\hline
 & Glasses case & Ramen & Pill bottle & Sandwich (upright) & Massage ball \\ \hline
Ours & 4/6 & 6/6 & 5/6 & 6/6 & 4/6 \\ \hline
Hardcoded & 3/3 & 3/3 & 1/3 & 3/3 & 3/3 \\ \hline
 & Castle & Tetrahedron & Hairspray bottle & Pear & Alcohol bottle \\ \hline
Ours & 6/6 & 1/6 & 6/6 & 5/6 & 6/6 \\ \hline
Hardcoded & 0/3 & 0/3 & 0/3 & 0/3 & 1/3 \\ \hline
 & Condiment bottle & Sandwich (side) & Lego & \textbf{Overall} & \textbf{Success \%} \\ \hline
Ours & 4/6 & 6/6 & 5/6 & \textbf{64/78} & \textbf{82.1} \\ \hline
Hardcoded & 1/3 & 0/3 & 0/3 & \textbf{15/39} & \textbf{38.5} \\ \hline
\end{tabular}
\label{tab:novel_obj}
\end{table}

\subsection{Quantitative Evaluation of Physical Constraints}
We provide quantitative physical constraint evaluations on grasps planned by our method. The evaluations are reported for each of the object categories. Table~\ref{tab:kinematic_stats} summarizes the signed distance (Equation~\eqref{eqn:signed_distance}) between the fingertips to the fitted object meshes. The distances should satisfy $D(\bm{p}, \bm{O})\in [d_{min}, d_{max}] = \left[-0.68, -0.32\right]$. Table~\ref{tab:dynamic_stats} summarizes the force and torque ratios, which should both be zero under wrench closure.

\begin{table}[h]
\caption{Evaluation of finger-object distance constraint on grasps planned by our method.}
\centering\scriptsize
\begin{tabular}{|c|c|c|c|c|}
\hline
 &  Median & Min. & Max. & Max violation \\ \hline
Seen & -0.32 & -0.68 & -0.24 & 0.08 \\ \hline
Familiar & -0.32 & -0.68 & -0.32 & 0.00 \\ \hline
Novel & -0.32 & -0.68 & -0.24 & 0.08 \\ \hline
Overall & -0.32 & -0.68 & -0.24 & 0.08 \\ \hline
\end{tabular}
\label{tab:kinematic_stats}
\end{table}

\begin{table}[h]
\caption{Force and torque ratios of the grasps generated by our method. All values are reported as \textit{median, (25th percentile, 75th percentile)}.}
\centering\scriptsize
\begin{tabular}{|c|c|c|}
\hline
 & Force ratio & Torque ratio \\ \hline
Seen & 0.00 (0.00, 0.01) & 0.01 (0.00, 0.27) \\ \hline
Familiar & 0.00 (0.00, 0.00) & 0.00 (0.00, 0.00) \\ \hline
Novel & 0.00 (0.00, 0.03) & 0.01 (0.00, 8.48) \\ \hline
Overall & 0.00 (0.00, 0.02) & 0.01 (0.00, 2.68) \\ \hline
\end{tabular}
\label{tab:dynamic_stats}
\end{table}

% Pre-rebuttal data
% \begin{table}[]
% \centering\scriptsize
% \caption{Seen objects.}
% \label{tab:seen_obj}
% \begin{tabular}{@{}l|cc|cc|cc|c@{}}
% \toprule
% Method / Object & \multicolumn{2}{c|}{Sugar Box} & \multicolumn{2}{c|}{Soup Can} & \multicolumn{2}{c|}{Mustard Bottle} & Overall \\
% Pose & Upright & Side & Upright & Side & Upright & Side & \multicolumn{1}{l}{} \\ \midrule
% Ours & 66.67\% & 100.00\% & 66.67\% & 33.33\% & 66.67\% & 66.67\% & 66.67\% \\
% CVAE only & 33.33\% & 33.33\% & 33.33\% & 33.33\% & 33.33\% & 0.00\% & 27.78\% \\
% Hardcoded & 100.00\% & 100.00\% & 66.67\% & 100.00\% & 100.00\% & 100.00\% & 94.44\% \\ \bottomrule
% \end{tabular}
% \end{table}

% \begin{table}[]
% \centering\scriptsize
% \caption{Familiar objects.}
% \label{tab:familiar_obj}
% \begin{tabular}{@{}l|cc|cc|cc|cc|c@{}}
% \toprule
% Method / Object & \multicolumn{2}{c|}{Webcam Box} & \multicolumn{2}{c|}{Mask Box} & \multicolumn{2}{c|}{Chips Can} & \multicolumn{2}{c|}{Soda Can} & Overall \\
% Pose & Upright & Side & Upright & Side & Upright & Side & Upright & Side & \multicolumn{1}{l}{} \\ \midrule
% Ours & \multicolumn{1}{r}{66.67\%} & \multicolumn{1}{r|}{33.33\%} & \multicolumn{1}{r}{66.67\%} & \multicolumn{1}{r|}{66.67\%} & \multicolumn{1}{r}{66.67\%} & \multicolumn{1}{r|}{66.67\%} & 100.00\% & 100.00\% & 70.83\% \\ \bottomrule
% \end{tabular}
% \end{table}

\subsection{Timing Statistics}
In Table~\ref{tab:timing}, we report the runtime and repeats of each stage in our pipeline during grasp planning. 

\begin{table}[h!]
\centering\scriptsize
\caption{Timing statistics of our method, shown as \textit{median, (25th percentile, 75th percentile)}.
}
\begin{tabular}{|c|c|c|c|c|c|c|c|}
\hline
 & CVAE time (s) & CVAE repeats & IK time (s) & IK repeats & BO time (s) & BO repeats & \textbf{Total Time (s)} \\ \hline
Seen & 0.95 (0.94, 0.97) & 2 (1, 3.75) & 13.02 (6.64, 18.88) & 2 (1, 4) & 1.99 (0.85, 2.90) & 1.5 (1, 2.75) & \textbf{14.99 (9.04, 23.60)} \\ \hline
Familiar & 0.95 (0.93, 0.97) & 2 (1, 4) & 10.91 (3.91, 21.52) & 2 (1, 4) & 0.78 (0.50, 2.25) & 1.5 (1, 2.25) & \textbf{13.73 (5.49, 31.68)} \\ \hline
Novel & 0.95 (0.93, 0.97) & 2 (1, 4) & 8.07 (1.47, 20.55) & 2 (1,4) & 2.61 (1.05, 5.35) & 2 (1, 3) & \textbf{14.40 (5.59, 34.85)} \\ \hline
\textbf{Overall} & \textbf{0.95 (0.93, 0.97)} & \textbf{2 (1, 4)} & \textbf{9.81 (2.16, 20.07)} & \textbf{2 (1,4)} & \textbf{2.22 (0.82, 4.03)} & \textbf{2 (1,3)} & \textbf{14.40 (5.67, 34.71)} \\ \hline
\end{tabular}
\label{tab:timing}
\end{table}

\bibliography{biblio}